\documentclass[a4paper,fleqn]{cas-sc}

\usepackage{algorithm}
\usepackage{algorithmic}
\usepackage{amsmath}

\usepackage[numbers]{natbib}
\usepackage{graphicx}

\newcommand{\cofirstSym}{\raisebox{0.1ex}{\scalebox{1.5}{\textdagger}}}
\newcommand{\corSym}{\raisebox{0.1ex}{\scalebox{1.5}{\textbf{*}}}}

\newcommand{\cofirstMark}{\textsuperscript{\scalebox{1.5}{\textdagger}}}
\newcommand{\corMark}{\textsuperscript{\scalebox{1.5}{\textbf{*}}}}

\usepackage{comment}

\usepackage{textcomp} % for \textdagger

\def\tsc#1{\csdef{#1}{\textsc{\lowercase{#1}}\xspace}}
\tsc{WGM}
\tsc{QE}
\tsc{EP}
\tsc{PMS}
\tsc{BEC}
\tsc{DE}

\def\name{MMDrive}

\usepackage{xcolor}
\definecolor{darkgreen}{RGB}{0,100,0}
\definecolor{lightgreen}{RGB}{144,238,144}
\definecolor{lightred}{RGB}{255,182,193}
\usepackage{booktabs}
\usepackage{multirow}
\usepackage[normalem]{ulem}
\usepackage{colortbl}

\usepackage{textcomp} % for \textdagger

\ExplSyntaxOn
\cs_gset:Npn \__first_footerline: { }
\ExplSyntaxOff
\begin{document}
\let\WriteBookmarks\relax
\def\floatpagepagefraction{1}
\def\textpagefraction{.001}

% Short title
\shorttitle{}

% Short author
\shortauthors{}

% Main title of the paper
\title [mode = title]{\name{}: Interactive Scene Understanding Beyond Vision with Multi-representational Fusion}   

% Title footnote mark
% eg: \tnotemark[1]
% \tnotemark[1]

% Title footnote 1.
% eg: \tnotetext[1]{Title footnote text}
% \tnotetext[<tnote number>]{<tnote text>} 
% \tnotetext[1]{This document is the results of the research
   % project funded by the National Science Foundation.}

% \tnotetext[2]{The second title footnote which is a longer text matter
%    to fill through the whole text width and overflow into
%    another line in the footnotes area of the first page.}

% First author
%
% Options: Use if required
% eg: \author[1,3]{Author Name}[type=editor,
%       style=chinese,
%       auid=000,
%       bioid=1,
%       prefix=Sir,
%       orcid=0000-0000-0000-0000,
%       facebook=<facebook id>,
%       twitter=<twitter id>,
%       linkedin=<linkedin id>,
%       gplus=<gplus id>]
\author[1]{Minghui Hou\cofirstMark}[style=chinese]

% Corresponding author indication
% \cormark[1]

% Footnote of the first author
% \fnmark[1]

% Email id of the first author
\ead{houmh21@mails.jlu.edu.cn}

% URL of the first author
% \ead[url]{www.cvr.cc, cvr@sayahna.org}

%  Credit authorship
% \credit{Writing – original draft, Visualization, Validation, Methodology, Formal analysis, Conceptualization.}

% Address/affiliation
\affiliation[1]{organization={College of Computer Science and Technology, Jilin University},
    % addressline={}, 
    city={Changchun},
    % citysep={}, % Uncomment if no comma needed between city and postcode
    % postcode={}, 
    % state={Jilin},
    country={China}}

% Second author
\author[2]{Wei-Hsing Huang\cofirstMark}[style=chinese]
\ead{whuang386@gatech.edu}
% \ead[URL]{www.sayahna.org}
% \credit{Data curation, Writing - Original draft preparation}
% Address/affiliation
\affiliation[2]{organization={Georgia Institute of Technology},
    % addressline={}, 
    city={Atlanta},
    % citysep={}, % Uncomment if no comma needed between city and postcode
    % postcode={}, 
    % state={GA},
    country={USA}}

% Third author
\author%
[3]
{Shaofeng Liang}[style=chinese]
% \cormark[2]
% \fnmark[1,3]
\ead{s23070053@s.upc.edu.cn}
% \ead[URL]{www.stmdocs.in}

\affiliation[3]{organization={Qingdao Institute of Software, College of Computer Science and Technology, China University of Petroleum (East China)},
    % addressline={Mepukada}, 
    city={Qingdao},
    % citysep={}, % Uncomment if no comma needed between city and postcode
    % postcode={695571}, 
    % state={GuangDong},
    country={China}}

% Forth author
\author%
[4]
{Daizong Liu}[style=chinese]
% \cormark[2]
% \fnmark[1,3]
\ead{sdaizongliu@whu.edu.cn}
% \ead[URL]{www.stmdocs.in}

\affiliation[4]{organization={Institute for Math \& AI, Wuhan University},
    % addressline={Mepukada}, 
    city={Wuhan},
    % citysep={}, % Uncomment if no comma needed between city and postcode
    % postcode={695571}, 
    % state={GuangDong},
    country={China}}

% Forth author
\author%
[5]
{Tai-Hao Wen}[style=chinese]
% \cormark[2]
% \fnmark[1,3]
\ead{taihaow@umich.edu}
% \ead[URL]{www.stmdocs.in}

\affiliation[5]{organization={University of Michigan, Ann Arbor},
    % addressline={Mepukada}, 
    % city={Wuhan},
    % citysep={}, % Uncomment if no comma needed between city and postcode
    % postcode={695571}, 
    % state={GuangDong},
    country={USA}}

% Fifth author
\author[1]{Gang Wang\corMark}[style=chinese]
\ead{gangwang@jlu.edu.cn}
% \ead[URL]{www.stmdocs.in}

% \affiliation[3]{organization={College of Computer Science and Technology, Jilin University},
%     % addressline={Mepukada}, 
%     city={Jilin},
%     % citysep={}, % Uncomment if no comma needed between city and postcode
%     % postcode={695571}, 
%     % state={GuangDong},
%     country={China}}

% Fifth author
\author[6]{Runwei Guan\corMark}[style=chinese]
\ead{runwayrwguan@hkust-gz.edu.cn}
% \ead[URL]{www.stmdocs.in}

\affiliation[6]{organization={Thrust of Artificial Intelligence, Hong Kong University of Science and Technology (Guangzhou)},
%     % addressline={Mepukada}, 
    city={Guangzhou},
%     % citysep={}, % Uncomment if no comma needed between city and postcode
%     % postcode={695571}, 
    % state={GuangDong},
    country={China}}

% Fifth author
\author%
[7]
{Weiping Ding}[style=chinese]
% \cormark[1]
% \fnmark[1,3]
\ead{ding.wp@ntu.edu.cn}
% \ead[URL]{www.stmdocs.in}

\affiliation[7]{organization={School of Artificial Intelligence and Computer Science, Nantong University},
    % addressline={Mepukada}, 
    city={Nantong},
    % citysep={}, % Uncomment if no comma needed between city and postcode
    % postcode={695571}, 
    % state={GuangDong},
    country={China}}
%\AddToHook{cmd/stmaddress/after}{%
%  \par\smallskip
%  \noindent
%  \cofirstSym\ Co-first author.\quad
%  \corSym\ Corresponding author.%
%}
\nonumnote{%
  \cofirstSym\ Co-first authors (equal contribution).\quad
  \corSym\ Corresponding authors.%
}

%% 共同一作和共同通讯的标识；
%\nonumnote{%
%  % \cofirstMark\ Co-first author.\quad
%  \corMark\ Corresponding author.%
%}
%\fntext[fn1]{Co-first author}
%\cortext[cor1]{Corresponding author}
% \cortext[cor2]{Principal corresponding author}

% Footnote text
%\fntext[fn1]{Co-first author}
% \fntext[fn2]{Another author footnote, this is a very long footnote and
%   it should be a really long footnote. But this footnote is not yet
%   sufficiently long enough to make two lines of footnote text.}

% For a title note without a number/mark
% \nonumnote{This note has no numbers. In this work we demonstrate $a_b$
%   the formation Y\_1 of a new type of polariton on the interface
%   between a cuprous oxide slab and a polystyrene micro-sphere placed
%   on the slab.
%   }

% Here goes the abstract
\begin{abstract}
Vision-language models enable the understanding and reasoning of complex traffic scenarios through multi-source information fusion, establishing it as a core technology for autonomous driving.
However, existing vision-language models are constrained by the image understanding paradigm in 2D plane, which restricts their capability to perceive 3D spatial information and perform deep semantic fusion, resulting in suboptimal performance in complex autonomous driving environments.
This study proposes \name{}, an multimodal vision-language model framework that extends traditional image understanding to a generalized 3D scene understanding framework. \name{} incorporates three complementary modalities, including occupancy maps, LiDAR point clouds, and textual scene descriptions. To this end, it introduces two novel components for adaptive cross-modal fusion and key information extraction. Specifically, the Text-oriented Multimodal Modulator dynamically weights the contributions of each modality based on the semantic cues in the question, guiding context-aware feature integration. The Cross-Modal Abstractor employs learnable abstract tokens to generate compact, cross-modal summaries that highlight key regions and essential semantics. 
Comprehensive evaluations on the DriveLM and NuScenes-QA benchmarks demonstrate that MMDrive achieves significant performance gains over existing vision-language models for autonomous driving, with a BLEU-4 score of 54.56 and METEOR of 41.78 on DriveLM, and an accuracy score of 62.7\% on NuScenes-QA.
\name{} effectively breaks the traditional image-only understanding barrier, enabling robust multimodal reasoning in complex driving environments and providing a new foundation for interpretable autonomous driving scene understanding.
\end{abstract}

% Use if graphical abstract is present
% \begin{graphicalabstract}
% \includegraphics{figs/grabs.pdf}
% \end{graphicalabstract}

% Research highlights
%\begin{highlights}
%\item A multimodal vision–language model that %enhances scene understanding beyond images.
%\item A text-oriented modulator to adjust modality %contributions and guide multimodal fusion.
%\item A cross-modal abstractor to distill key %information and generate compact representations.
%\end{highlights}

% Keywords
% Each keyword is seperated by \sep

\maketitle

\section{Introduction}
\label{sec:introduction}

% figure1_intro
\begin{figure}
    \begin{center}
        \includegraphics[width=0.9\linewidth]{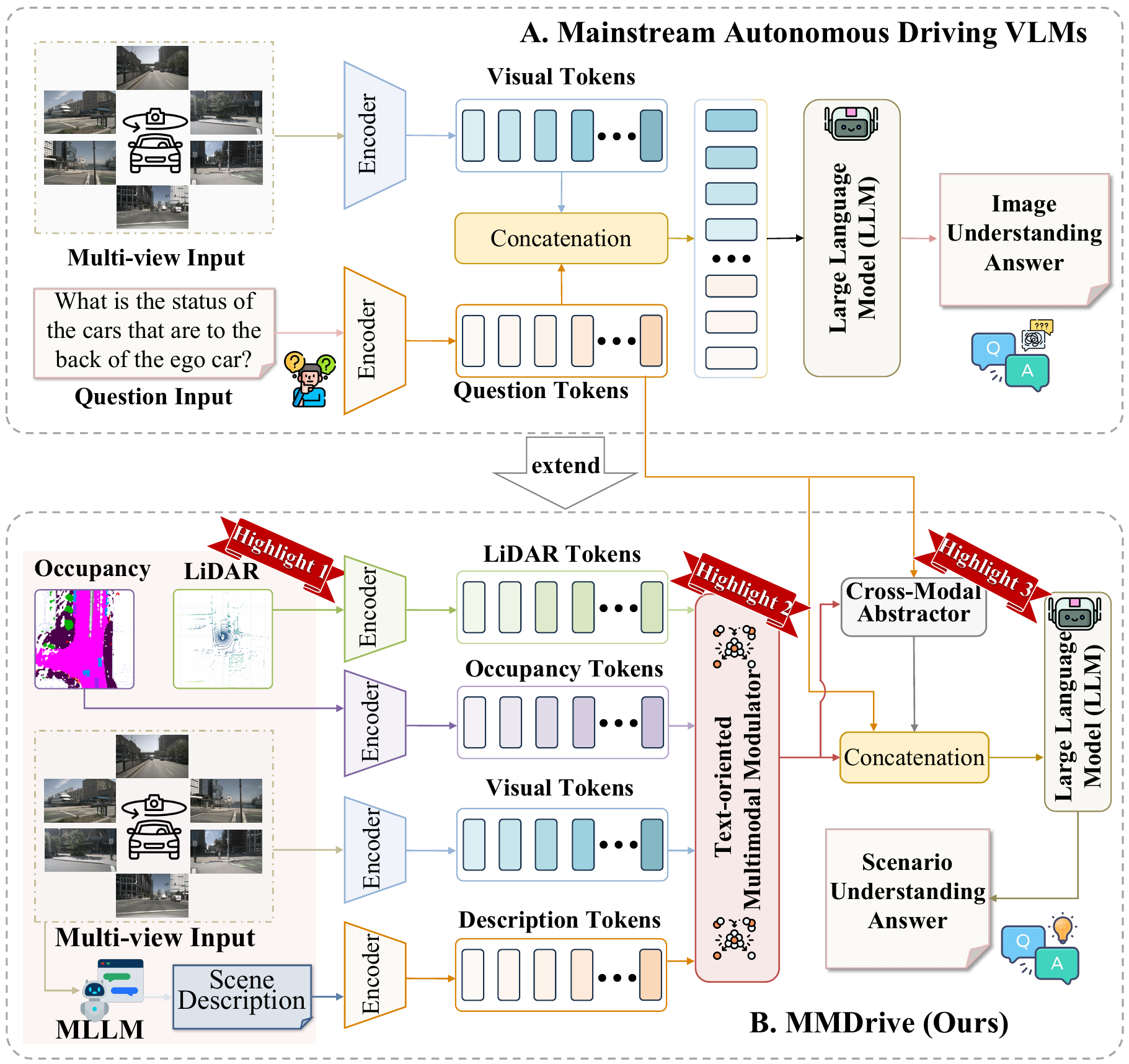}
    \end{center}
    % \vspace{-0.5cm}
\caption{
Comparison between mainstream VLMs for autonomous driving and the proposed \name{}.
(A) Mainstream Image Understanding Paradigm: Image and text features are extracted encoders and combined through projection, limiting cross-modal interaction.
(B) \textbf{\name{}}: Our framework incorporates occupancy, LiDAR, and scene description modalities, extending the conventional image understanding paradigm toward holistic scene understanding. It also incorporates the TMM and CMA to enable multimodal information fusion, thereby enhancing representational capability and adaptability in complex driving scenarios.
}
\label{fig_intro}
\end{figure}

Driven by rapid advances in multi-source information fusion~\cite{danish2026a}, Vision-Language Models (VLMs) have demonstrated remarkable capabilities in Visual Question Answering (VQA), enabling their extensive application within autonomous driving systems to enhance scene perception and decision support~\cite{sima2024drivelm}. 
As an emerging multimodal task, VLMs for autonomous driving aim to perceive and understand complex driving environments through VQA. Unlike traditional approaches~\cite{hou2025polarbevu}, VLMs not only enhance perceptual capabilities but also improve interpretability and enable semantic reasoning for decision support~\cite{sapkota2026object}. These advantages establish strong foundations for the safety and robustness of autonomous driving, thereby positioning VLMs as one of the most promising research directions in the field~\cite{li2025generative}.

Current mainstream VLMs for autonomous driving~\cite{Gopalkrishnan2024emvlm4ad,zhang2025mpdrive}, as shown in Figure~\ref{fig_intro}(A), adopt a dual-branch ``image understanding'' paradigm, where the vision encoder extracts visual features and the language encoder encodes textual questions. Adapters map both modalities into the Large Language Models (LLMs) token space, where they are concatenated and decoded to produce the final answers.
Representative works include DriveLM-Agent~\cite{sima2024drivelm}, which employs a graph-based VQA model to capture logical dependencies across driving stages and facilitate human-like multi-step reasoning. 
However, it relies heavily on structured representations and lacks efficient mechanisms for multi-view information integration.
To address this limitation, EM-VLM4AD~\cite{Gopalkrishnan2024emvlm4ad} introduces a multi-frame embedding strategy that integrates information from multiple views and employs a lightweight model to enhance computational efficiency. 
Building upon this, MiniDrive~\cite{zhang2024minidrive} further improves 2D feature processing by employing multi-level token embeddings and incorporating feature engineering mixture of experts units to improve multi-image processing performance.
Nevertheless, these methods primarily focus on representation enhancement rather than adaptive reasoning.
LaVida Drive~\cite{jiao2024lavida} enhances semantic understanding by integrating a query aware dynamic selection mechanism and spatial temporal enhancement modules, improving the integration of spatial and temporal information for more effective VQA performance.
More recently, MPDrive~\cite{zhang2025mpdrive} transforms the generation of complex spatial coordinate generation into text-based visual token prediction, thereby improving linguistic consistency and accuracy in spatial representation.

% figure1_intro
\begin{figure}
    \begin{center}
        \includegraphics[width=0.9\linewidth]{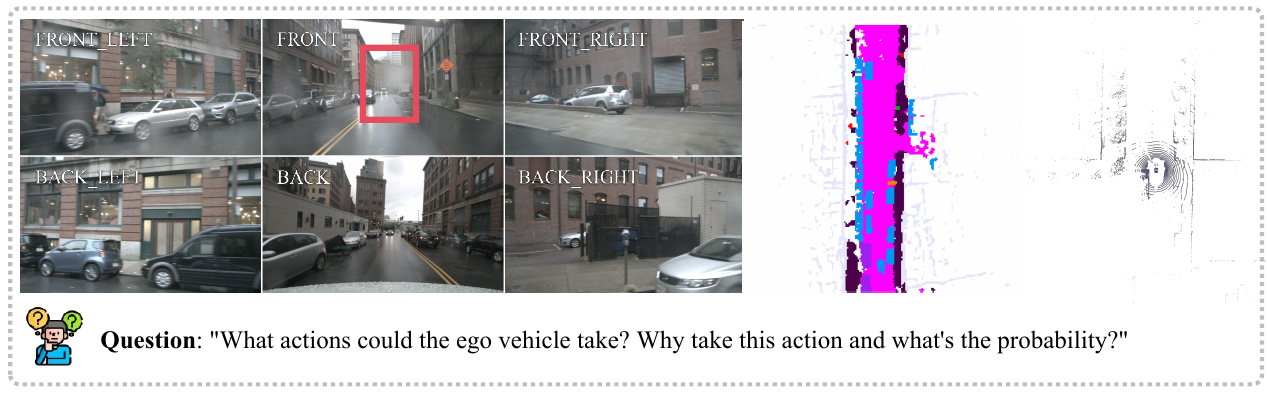}
    \end{center}
    % \vspace{-0.5cm}
\caption{
Image-only sensing poses difficulties for object recognition in complex autonomous driving scenarios.
}
\label{fig:intro_sample}
\end{figure}

Despite their significant contributions, existing methods still adhere to the traditional ``image understanding'' paradigm established in general VQA tasks. Moreover, autonomous driving scenarios are dynamic and complex by nature, and traditional methods relying solely on 2D visual representations lack the essential 3D spatial information and depth perception required for effective autonomous driving. Consequently, they are unable to meet the requirements of VQA tasks for autonomous driving, particularly in terms of precise spatial understanding and dynamic interaction.
%
% As illustrated in Figure~\ref{fig:intro_sample}, relying solely on image data is inadequate for reliably identifying objects that are highly similar to the background or partially occluded.
As shown in Figure~\ref {fig:intro_sample}, the perception system’s forward-facing camera is occluded, and the driving scenario is complex.In this case, relying solely on image data fails to enable accurate environmental perception and reliable scene understanding.
Moreover, visual information remains at the perception level, primarily characterizing sensory or metric attributes of the physical world. In contrast, VQA tasks for autonomous driving require higher-level comprehension and reasoning over complex dynamic objects and semantic associations within a scene. This demand for intricate semantic mappings from limited training data results in a significant cognitive gap between perception and understanding.
Unlike existing image-only paradigms, this work's core insight is that incorporating multimodal information provides an effective approach to bridging this gap. 
However, current multimodal fusion methods still faces two major challenges. First, different textual queries focus on distinct modalities. For instance, some questions emphasize depth information, whereas others concern spatial layout or 2D visual cues. Conventional fusion strategies that simply concatenate multimodal features tend to overlook these semantic distinctions, thereby weakening the model's focus on modality-specific features.
Second, in dynamic and highly complex environments, the model struggles to efficiently prioritize information within a vast multimodal space, making it difficult for LLMs to attend to critical regions and key semantic cues.

To address the aforementioned limitations, this work proposes \textit{\name{}}, an end-to-end multimodal vision-language model (VLM) for autonomous driving, as illustrated in Figure~\ref{fig_intro}(B). 
\name{} extends the conventional image understanding paradigm into a generalized scene understanding paradigm to enable deeper semantic reasoning. To this end, unlike previous purely image-based approaches, it integrates occupancy, LiDAR, and scene description modalities, which provide complementary spatial, depth, and semantic cues to enhance scene understanding.
Specifically, the occupancy features effectively capture spatial distributions and provide dense 3D structural information for scene understanding~\cite{xu2025occsuevey}. The LiDAR modality serves as an explicit depth complement. Furthermore, scene descriptions are generated through a VLM and a large language model (LLM) via a carefully designed two-stage prompting strategy, thereby enhancing the model’s semantic understanding of driving scenes.
\name{} overcomes the limitations of the image understanding paradigm through two complementary modules that enable adaptive multimodal fusion and key information extraction.
The \textit{Text-oriented Multimodal Modulator} (TMM) dynamically adjusts the importance of multiple modalities during information fusion according to the semantic characteristics of textual queries.
The \textit{Cross-Modal Abstractor} (CMA) adopts a two-stage design ``text comprehension and multimodal content extraction'' to generate cross modal abstractions, enabling the LLM to more effectively attend to critical regions and salient information.
Together, these components collaboratively enable robust multimodal fusion and efficient reasoning in complex driving environments.
The effectiveness of \name{} is validated on the DriveLM~\cite{sima2024drivelm} and NuScenes-QA~\cite{qian2024nuscenesqa} benchmarks. The main contributions of this work are summarized as follows:

\begin{itemize}
    \item The Text-oriented Multimodal Modulator (TMM), a module designed to establish a dynamic association mechanism between text query semantics and modality contribution. It adjusts the weights of each modality based on the semantic features of textual queries, ensuring precise alignment of the multimodal fusion process with the query intent.

    \item The Cross-Modal Abstractor (CMA), a module proposed to extract essential information for scene understanding, employs learnable abstract tokens to generate a compact cross-modal abstraction, refining key information from fused multimodal representations.

    \item An end-to-end autonomous driving VLM, \name{}, is designed by integrating TMM and CMA within a unified framework. It incorporates occupancy, LiDAR, and scene description modalities, extending the conventional image understanding paradigm toward comprehensive scene understanding. Extensive experiments on DriveLM~\cite{sima2024drivelm} and NuScenes-QA~\cite{qian2024nuscenesqa} benchmarks demonstrate the effectiveness and superiority of \name{}.
\end{itemize}

The remainder of this paper is organized as follows. Section \ref{sec:related_work} provides a review of related work in vision-language model and its applications in autonomous driving. Section \ref{sec:methods} details the proposed \name{} methodology and components. Section \ref{sec:experiments} presents the experimental setup and results, and Section \ref{sec:conclusions} concludes the paper with key findings and explores future research directions.

\section{Related Works}
\label{sec:related_work}

\subsection{Vision-Language Foundation Models}
Modern VLMs are based on Transformer-style backbones for long-range dependency modeling, together with large-scale pretraining on text and images~\cite{Radford21Learning, Zhang24Vision, zhang2026cost}. Early cross-modal encoders typically followed either a dual-stream pathway or a single-stream pathway, where dual-stream approaches use separate visual and textual encoders with late interaction and single-stream approaches jointly contextualize visual tokens and words within a unified encoder~\cite{Zhang24Vision}. CLIP~\cite{Radford21Learning} established strong open-vocabulary grounding through scaled contrastive alignment on noisy web image-text pairs. Building on this, BLIP~\cite{Li22BLIP} improves web‑scale pretraining quality and unifies vision-language understanding and generation, and BLIP‑2~\cite{Li23BLIP2} efficiently couples frozen vision encoders with frozen LLMs. Decoder-only LLMs equipped with visual adapters and instruction tuning, such as LLaVA~\cite{Liu23Visual}, InstructBLIP~\cite{Dai23InstructBLIP}, and MiniGPT‑4~\cite{Zhu24MiniGPT} expand few-shot reasoning and conversational competence. As a stress test for compositional reasoning, GQA~\cite{Hudson19GQA} standardized protocols for measuring cross-modal understanding and error patterns. 
Recent surveys also review how VLMs are repurposed across vision tasks, offering a backdrop for domain-specific adaptations~\cite{Zhang24Vision, liu2026embracing}. Beyond alignment‑first encoders and instruction‑tuned decoders, a parallel research direction explores unified generative sequence‑to‑sequence interfaces for both multimodal and unimodal tasks, where captioning, grounding, question answering, classification, and even image generation share a common input-output format. This consolidation is orthogonal to contrastive pretraining and provides a practical route to simplify pretraining/finetuning pipelines while maintaining broad coverage across benchmarks. Existing methods rely on fixed vision-language alignment, enhancing performance through increased pretraining scale or adjusted model architecture. However, their reasoning capabilities remain insufficient in complex scenarios. \name{} facilitates efficient multimodal information fusion, significantly improving the model's understanding and reasoning abilities.

\subsection{Vision-Language Models for Autonomous Driving}
Autonomous driving shifts VLMs from single-image perception toward scene-level reasoning across time, viewpoints, and modalities. 
New datasets extend beyond 2D images to emphasize 3D awareness, long-horizon context, and rule-centric competence: nuScenes‑QA supplies multi-modal, 3D-aware QA at scale ~\cite{qian2024nuscenesqa}; LingoQA~\cite{Marcu24LingoQA} focuses on video-centric QA; Passing the Driving Knowledge Test~\cite{wei2025passing} reframes evaluation around traffic rules; and OmniDrive~\cite{Wang25OmniDrive} explores holistic scene-level vision-language understanding with counterfactual reasoning over diverse driving scenarios. VLADBench~\cite{li2025fine} introduces a fine-grained benchmark with close-form QAs that progress from foundational traffic knowledge and elements to advanced reasoning for ego decision-making and planning in autonomous driving. DriveLM~\cite{sima2024drivelm} formulates driving QA over a perception, prediction and planning graph with explicit inter-question dependencies. EM‑VLM4AD~\cite{Gopalkrishnan2024emvlm4ad} proposes lightweight, gated pooling attention over multi‑frame embeddings. MiniDrive~\cite{zhang2024minidrive} maps multi-level 2D features to text tokens with FE‑MoE and dynamic instruction adapters; LaVida Drive~\cite{jiao2024lavida} selects and restores query-aware spatio‑temporal tokens to retain resolution-critical details; MPDrive~\cite{zhang2025mpdrive} replaces ad‑hoc coordinate strings with marker-based visual prompting for stronger spatial grounding. A critical requirement is geometry-aware fusion. Unified BEV spaces align camera and LiDAR into a scene-centric representation, and occupancy-style 3D representations provide voxel‑level context and improve robustness to occlusion and layout ambiguity~\cite{xu2025occsuevey, Liu23BEVFusion}. Language-guided stacks complement pure QA by linking description, analysis, and planning: LMDrive~\cite{Shao24LMDrive} demonstrates closed-loop control with language supervision; DriveGPT‑4~\cite{Xu24DriveGPT4} injects interpretable statements into end‑to‑end control; Dolphins~\cite{Ma24Dolphins} emphasizes grounded multi-frame reasoning; DriveVLM~\cite{Tian24DriveVLM} integrates scene description, scene analysis, and hierarchical planning; Reason2Drive~\cite{Nie24Reason2Drive} builds chain-based reasoning corpora; SimLingo~\cite{Renz25SimLingo} aligns vision-only control with language actions; Language Prompt for autonomous driving~\cite{Wu25Language} bridges perception and trajectory generation; and LiDAR‑LLM~\cite{Yang25LiDAR} treats raw LiDAR as a first-class modality for language alignment. SOLVE~\cite{chen2025solve} synergizes VLMs with end-to-end planners via feature-level knowledge sharing through a shared visual encoder, and proposes a Trajectory Chain-of-Thought paradigm with temporal decoupling for efficient cooperation.
VLR-Driver~\cite{kong2025vlr} proposes a multi-modal vision-language-reasoning framework with spatiotemporal Chain-of-Thought to analyze safety risks and other agents' intentions, and constructs a multi-modal reasoning-decision dataset with closed-loop validation in CARLA. 
ReasonDrive~\cite{Chahe25ReasonDrive} tailored for driving QA further indicate that targeted supervision can improve both accuracy and yield more transparent reasoning traces. Despite the significant contributions of the aforementioned methods, they adhere to the general VLM paradigm. Relying solely on images for reasoning struggles to address the complexities of dynamic autonomous driving scenarios. \name{} facilitates more precise reasoning in complex environments through efficient multimodal fusion and abstract extraction mechanisms.

% \paragraph{Position of our work.} 
Existing methods largely inherit an image-understanding paradigm, which (i) limits the utilization of 3D information and semantic priors, and (ii) typically fuses modalities in manners unaligned with query semantics, neglecting the extraction of key information from multimodal data.
In contrast, \name{} adopts a scene-understanding paradigm by (i) incorporating occupancy, LiDAR, and scene descriptions, and (ii) introducing two cooperating modules: the TMM, which dynamically adjusts modality importance according to the query, and the CMA, which distills a compact cross-modal summary before decoding. 
Together they strengthen multimodal fusion and reasoning in complex driving environments.

\section{Methods}
\label{sec:methods}

 % figure2_overview
 \begin{figure}
     \begin{center}
         \includegraphics[width=0.99\linewidth]{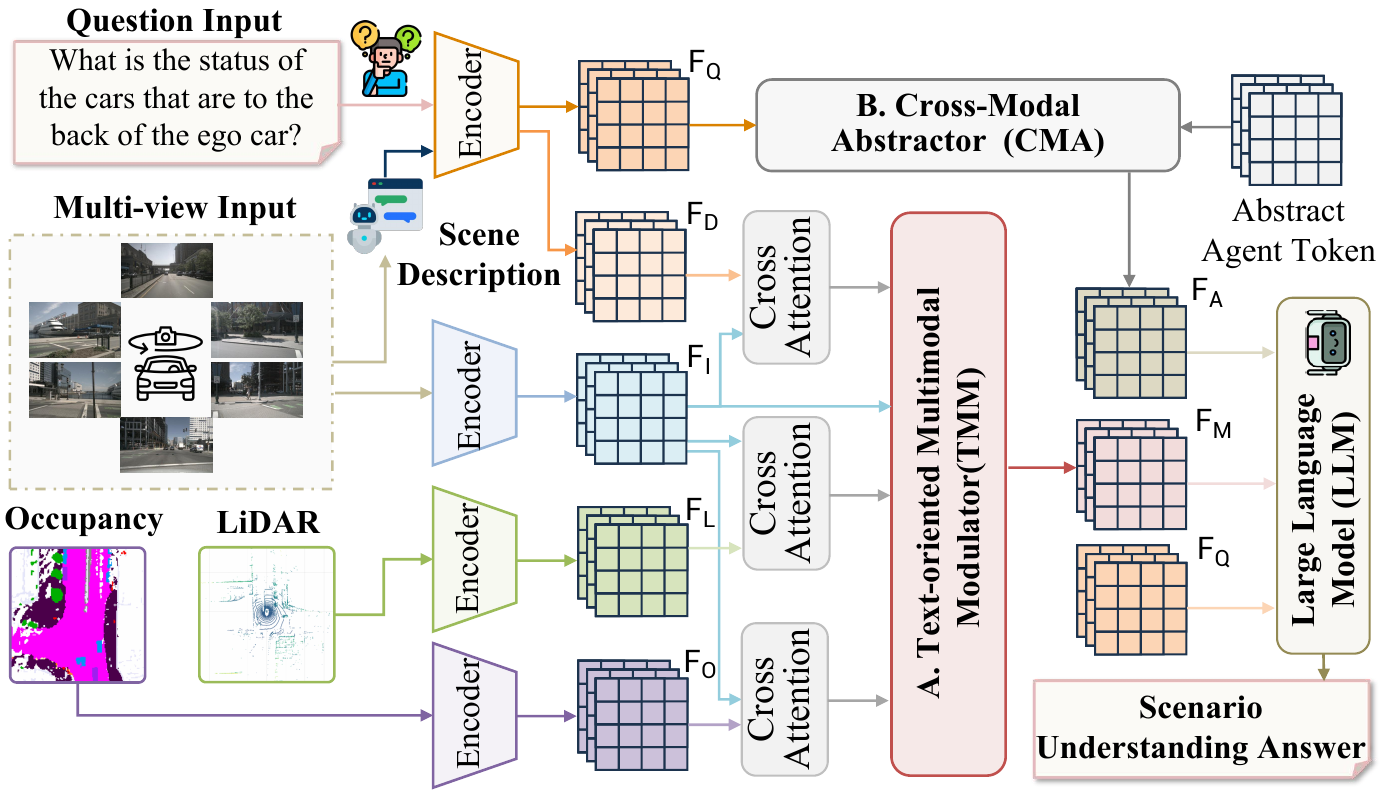}
     \end{center}
     % \vspace{-0.5cm}
 \caption{
Overview of the \name{} architecture. (1) The model takes multi-view images, text questions, occupancy, LiDAR, and scene descriptions as inputs. (2) It first employs frozen encoders to extract modality-specific features. (3) The Text-oriented Multimodal Modulator (TMM) dynamically adjusts the contribution of multimodal information based on the semantic content of text questions, achieving adaptive multimodal fusion. (4) The Cross-Modal Abstractor (CMA) further refines the fused multimodal representations by distilling critical information. (5) Finally, the fused representations are fed into a LLM to generate the final answer.
 }
 \label{fig_overview}
 \end{figure}

Existing VLMs largely follow the conventional image-understanding paradigm inherited from general VQA tasks. This approach proves insufficient for holistic scene comprehension in autonomous driving contexts, as it does not fully exploit multimodal cues. To overcome this shortcoming, we propose \name{}, a novel framework that generalizes the traditional ``image understanding'' paradigm into a unified ``scene understanding'' paradigm through the integration of heterogeneous modalities such as images, occupancy grids, depth maps, and textual scene descriptions. As depicted in Figure~\ref{fig_overview}, \name{} introduces two key components: the Text-oriented Multimodal Modulator (TMM) and the Cross-Modal Abstractor (CMA), designed to enable adaptive multimodal fusion and distill essential scene information. The TMM dynamically modulates the contribution of each modality based on the semantic cues in the question, thereby guiding the fusion of multimodal features. The CMA further refines scene semantics via learnable abstract tokens, enhancing the representation of cross-modal interactions. Together, these innovations allow \name{} to achieve more comprehensive and context-aware scene interpretation tailored to autonomous driving.

\subsection{Problem Formulation}
\label{subsec:problem_formulation}

The VLMs for autonomous driving aim to achieve comprehensive scene understanding by integrating multimodal sensory data, facilitating accurate perception and decision-making in complex driving environments. Formally, given a set of driving scene images $\mathcal{I} = \{\mathbf{I}_1, \mathbf{I}_2, \ldots, \mathbf{I}_{n}\}$, where $n$ denotes the number of images. The corresponding textual questions are denoted as $\mathcal{Q} = \{w_1, w_2, \ldots, w_{m}\}$, where $w_m$ represents lexical units and $m$ is the question length. The model's objective is to generate corresponding answers $\mathcal{A} = \{a_1, a_2, \ldots, a_{l}\}$. This task can be formulated as a conditional probability modeling problem:
 \begin{equation}
     \label{eq:task_definition}
         P(\mathcal{A} \mid \mathcal{I}, \mathcal{Q}; \theta) = \prod_{t=1}^{l} P(a_t \mid a_{<t}, \mathbf{F}_\mathrm{I}, \mathbf{F}_\mathrm{Q}; \theta),
 \end{equation}
where $\mathbf{F}_\mathrm{I}$ denotes the image features and $\mathbf{F}_\mathrm{Q}$ represents the question text features. $\theta$ denotes the model parameters, and $a_{<t} = \{a_1, \ldots, a_{t-1}\}$ represents the previously generated answer prefix.

However, the image-only understanding paradigm fails to meet the requirements of high complexity and dynamism inherent in autonomous driving scenarios. To address this challenge, \name{} introduces multimodal information $\mathcal{M}$, extending the traditional image understanding paradigm to a scene understanding paradigm. 
Specifically, to construct a comprehensive scene representation, our framework integrates three complementary modalities. Occupancy grids ($\mathcal{M}_{\mathrm{O}}$) are incorporated to deliver dense, probabilistic 3D spatial layout information, capturing the drivable and occupied regions of the environment. LiDAR point clouds ($\mathcal{M}_{\mathrm{L}}$) are introduced to supply precise, explicit geometric and depth cues, crucial for understanding object shapes and distances. Furthermore, we propose a novel two-stage generation strategy to produce rich scene descriptions ($\mathcal{M}_{\mathrm{D}}$), which encapsulate high-level semantic context and relational priors, thereby bridging the gap between low-level sensor data and abstract scene reasoning. This multi-faceted integration is fundamental to advancing from mere image analysis to holistic scene understanding.
Accordingly, the multimodal information fusion task for autonomous driving scene understanding can be formally defined as:
 \begin{equation}
 \label{eq:multimodal_task}
     P(\mathcal{A} \mid \mathcal{I}, \mathcal{M}, \mathcal{Q}; \theta) = \prod_{t=1}^{l} P(a_t \mid a_{<t}, \mathbf{F}_\mathrm{M}, \mathbf{F}_\mathrm{Q}; \theta),
 \end{equation}
where $\mathbf{F}_\mathrm{M}$ represents the fused multimodal representation including the original image information. The core objective of \name{} is to learn effective multimodal fusion strategies and information abstraction mechanisms, thereby enhancing the model's understanding and reasoning capabilities for complex driving scenes.

\subsection{Multimodal Information Encoding}
\label{subsec:multimodal_encoder}

To model comprehensive scene understanding, \name{} employs a multi-path encoder architecture that independently processes different modalities of input information. The design of each encoder is tailored to the specific characteristics of its corresponding modality, maximizing information extraction efficiency.

\textbf{Image Encoder}. To obtain high-quality image representations, \name{} employs UniRepLKNet-A~\cite{ding2024unireplknet} as the image encoder. This encoder leverages large-kernel convolution operations and incorporates a multi-scale feature pyramid structure, which adaptively adjusts kernel sizes to extract features at different scales while fusing multi-scale semantic and detailed information. Furthermore, UniRepLKNet-A integrates residual connections and attention mechanisms, effectively capturing long-range dependencies in images and enhancing the model's expressive capability. For an input image $\text{I} \in \mathbb{R}^{C \times H_\text{I} \times W_\text{I}}$, where $C$ denotes the number of channels, and $H_\text{I}$ and $W_\text{I}$ represent the image height and width, respectively, the image features are extracted as:
 \begin{equation}
     \mathbf{F}_\text{I} = \text{ImageEncoder}(\mathbf{I}),
 \end{equation}
where $\mathbf{F}_\text{I} \in \mathbb{R}^{S_\text{I} \times D_\text{I}}$, with $S_\text{I}$ representing the number of image feature tokens and $D_\text{I}$ denoting the feature embedding dimension. Image features from multiple viewpoints are encoded independently to preserve their respective spatial topologies. This independent encoding strategy ensures the integrity of multi-view image information, establishing a solid foundation for subsequent multimodal fusion and VQA.

\textbf{Question Text Encoder}. To efficiently capture rich semantic information and validate the superiority of the proposed module, \name{} employs the widely used T5~\cite{raffel2020t5} as the text encoder~\cite{Gopalkrishnan2024emvlm4ad, zhang2024minidrive,jiao2024lavida}. T5 features a unified architecture, supporting a wide range of natural language processing (NLP) tasks. It also accommodates large-scale pretraining paradigms and integrates multimodal data. This lays a solid technical foundation for emerging cross-modal research fields, including VLMs. For an input textual question $\mathcal{Q}$, the text is first tokenized through a tokenizer:
\begin{equation}
    \mathcal{W} = \text{Tokenizer}(\mathcal{Q}),
\end{equation}
where $\mathcal{W}$ represents the token sequence $\mathcal{W} = \{w_1, w_2, \ldots, w_{m}\}$, with $w_m$ denoting lexical units and $m$ indicating the question length. The features are then extracted as:
\begin{equation}
    \mathbf{F}_{\text{Q}} = \text{TextEncoder}(\mathcal{W}),
\end{equation}
where $\mathbf{F}_\text{Q} \in \mathbb{R}^{S_Q \times D_Q}$, with $S_Q$ representing the number of text feature tokens and $D_Q$ denoting the feature embedding dimension. Since different questions may have varying sequence lengths, T5 employs a padding mechanism during batch processing to align all questions within the same batch to a uniform length. This text encoder extracts deep semantic features from text, providing robust support for subsequent multimodal weight prediction.

\textbf{Occupancy Encoder}. Unlike traditional two-dimensional visual representations, occupancy grids can precisely capture the 3D spatial structure of scenes, making them effective for supplementing 3D spatial information. Specifically, this work employs a pre-trained generative framework~\cite{li2025uniscene} as the encoder, which is based on diffusion models and implements the denoising diffusion process through a Diffusion Transformer backbone, generating occupancy features $\mathbf{F}_{\mathrm{O}} \in \mathbb{R}^{S_{\mathrm{O}} \times D_{\mathrm{O}}}$, where $S_O$ denotes the number of tokens and $D_O$ represents the feature embedding dimension. These features encode the spatial occupancy probability distribution of autonomous driving scenes, effectively representing critical information such as spatial structure and object distribution, thereby providing dense 3D spatial priors for subsequent multimodal fusion modules.

 %figure_scene_description
 \begin{figure}
     \begin{center}
         \includegraphics[width=0.95\textwidth]{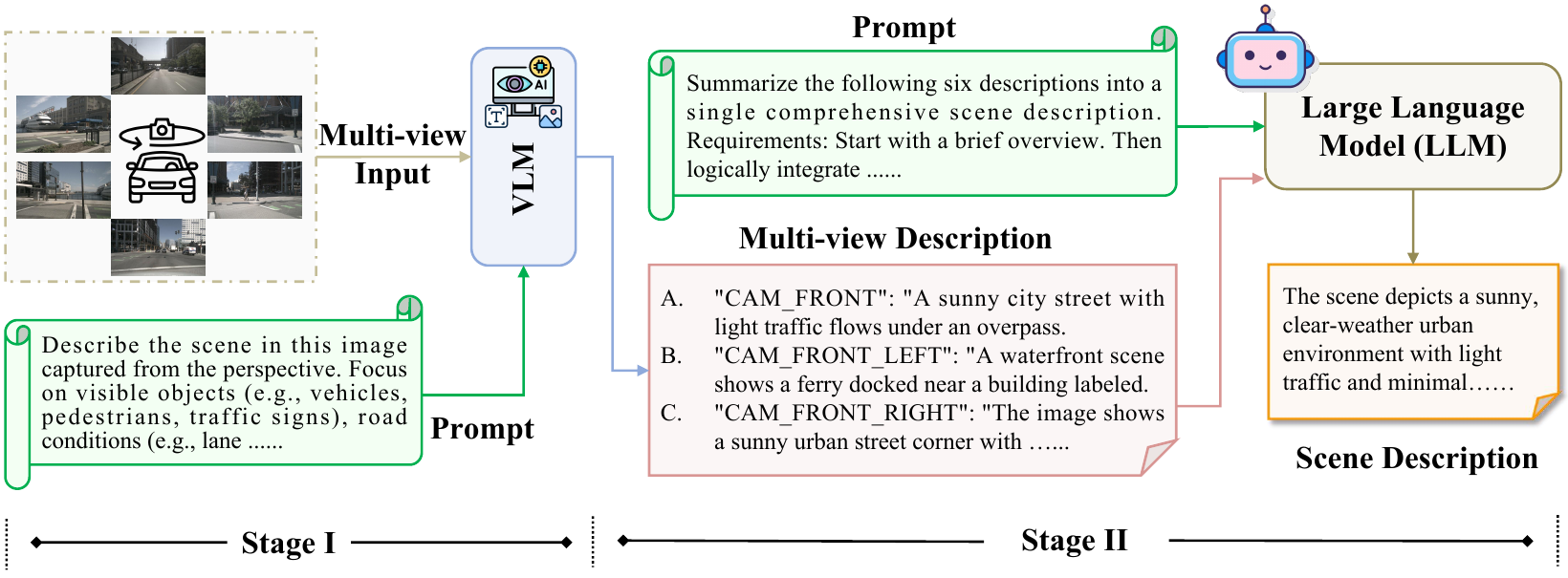}
     \end{center}
     % \vspace{-0.5cm}
 \caption{
Illustration of scene description generation through a two-stage hierarchical strategy. In the first stage, multi-view images and text prompts are fed into a Vision-Language Model (VLM) to generate corresponding multi-view descriptions. In the second stage, these multi-view descriptions are input to a Large Language Model (LLM) along with a summarization prompt to produce the final scene description.
 }
 \label{fig:scene_desc}
 \end{figure}

%8
\textbf{LiDAR Encoder}. While occupancy provides global spatial information, it remains limited in capturing local details and depth precision. To effectively model depth and geometric details of scenes, \name{} incorporates LiDAR information to explicitly supplement depth and geometric features. This work employs a pre-trained LiDAR encoder~\cite{zhao2023michel} to extract LiDAR features. Given a LiDAR point cloud $\mathcal{P} = \{\mathbf{p}_i\}_{i=1}^{N_{\mathrm{pts}}}$, where $\mathbf{p}_i \in \mathbb{R}^3$ represents 3D spatial coordinates, the normal vector $\mathbf{n}_i \in \mathbb{R}^3$ is first computed for each point to enhance local structural representation. 

This work adopts the classical local neighborhood covariance analysis method to estimate point cloud normals. Considering the non-uniform characteristic of point cloud data, which is dense nearby and sparse at distance, a neighborhood radius $r$ is defined for point $\mathbf{p}_i$, containing at most $k_{\max}$ nearest neighbors. Specifically, all points within radius $r$ are identified; if the count is smaller than $k_{\max}$, all points within this range are used for normal computation. Otherwise, the $k_{\max}$ nearest points are selected based on distance sorting:
\begin{equation}
    \mathcal{N}_r(i) = \{\mathbf{p}_j \mid \|\mathbf{p}_j - \mathbf{p}_i\|_2 \le r\}, \quad |\mathcal{N}_r(i)| \le k_{\max}.
\end{equation}

After obtaining the neighborhood points, the normal vector is computed through covariance analysis. First, the neighborhood mean $\boldsymbol{\mu}_i$ is calculated:
\begin{equation}
\boldsymbol{\mu}_i = \frac{1}{|\mathcal{N}_r(i)|} \sum_{\mathbf{p}_j \in \mathcal{N}_r(i)} \mathbf{p}_j.
\end{equation}
where $\mathcal{N}_r(i)$ denotes the $k$-nearest neighbor set of point $\mathbf{p}_i$.
Subsequently, the covariance matrix $\mathbf{C}_i$ is computed:
 \begin{equation}
     \mathbf{C}_i = \frac{1}{|\mathcal{N}_r(i)|} \sum_{\mathbf{p}_j \in \mathcal{N}_r(i)} (\mathbf{p}_j - \boldsymbol{\mu}_i)(\mathbf{p}_j - \boldsymbol{\mu}_i)^\top
 \end{equation}
where $\boldsymbol{\mu}_i$ is the neighborhood centroid.
 
The eigenvector corresponding to the smallest eigenvalue of the covariance matrix $\mathbf{C}_i$ is then extracted as the estimated normal vector $\mathbf{n}_i$. Subsequently, the original coordinates are concatenated with the normal vector to form $[\mathbf{p}_i; \mathbf{n}_i] \in \mathbb{R}^6$, which is then processed through the point cloud encoder to generate LiDAR features $\mathbf{F}_{\mathrm{L}} \in \mathbb{R}^{B \times N_{\mathrm{L}} \times D_{\mathrm{L}}}$, where $N_{\mathrm{L}}$ denotes the number of LiDAR tokens and $D_{\mathrm{L}}$ represents the feature dimension. These point cloud features provide explicit depth and geometric detail information for subsequent multimodal fusion modules.

\textbf{Scene Description Encoder}. To model high-quality semantic information of scenes, this work designs a two-stage generation strategy with carefully crafted prompts to generate scene descriptions from multi-view images, as illustrated in Figure~\ref{fig:scene_desc}. Specifically, in the first stage, a view-specific prompt $\mathcal{P}_{\mathrm{view}}^{(i)}$ is constructed for each viewpoint $\mathbf{I}_i$. This prompt, along with the corresponding image, is input to a pre-trained VLM~\cite{qwen25vl}, which generates a corresponding single-view description $\mathcal{D}_i$ for each viewpoint image:
\begin{equation}
\label{eq:view_desc}
    \mathcal{D}_i = \mathrm{VLM}(\mathbf{I}_i, \mathcal{P}_{\mathrm{view}}^{(i)}), \quad i \in \{1, \ldots, N_{\mathrm{view}}\}.
\end{equation}

In the second stage, all single-view descriptions are aggregated into a multi-view description set $\mathcal{D}_{\mathrm{multi}} = \{\mathcal{D}_1, \ldots, \mathcal{D}_{N_{\mathrm{view}}}\}$, and a scene-level prompt $\mathcal{P}_{\mathrm{scene}}$ is designed. Both are input to a LLM~\cite{qwen3} to generate the final scene description:
\begin{equation}
\label{eq:scene_desc}
    \mathcal{D}_{\mathrm{scene}} = \mathrm{LLM}(\mathcal{D}_{\mathrm{multi}}, \mathcal{P}_{\mathrm{scene}}).
\end{equation}
This two-stage hierarchical generation strategy preserves fine-grained information from multiple viewpoints while generating a unified global scene semantic representation. Finally, the scene description $\mathcal{D}_{\mathrm{scene}}$ is encoded into feature representations $\mathbf{F}_{\mathrm{D}} \in \mathbb{R}^{S_{\mathrm{D}} \times D_{\mathrm{D}}}$ through a pre-trained LLM~\cite{raffel2020t5}, where $S_{\mathrm{D}}$ denotes the sequence length of the scene description and $D_{\mathrm{D}}$ represents the feature embedding dimension, providing high-quality semantic priors for subsequent multimodal fusion.

 \subsection{Text-oriented Multimodal Modulator}
 \label{subsec:tmm}

% figure_tmm
 \begin{figure}
     \begin{center}
         \includegraphics[width=0.95\linewidth]{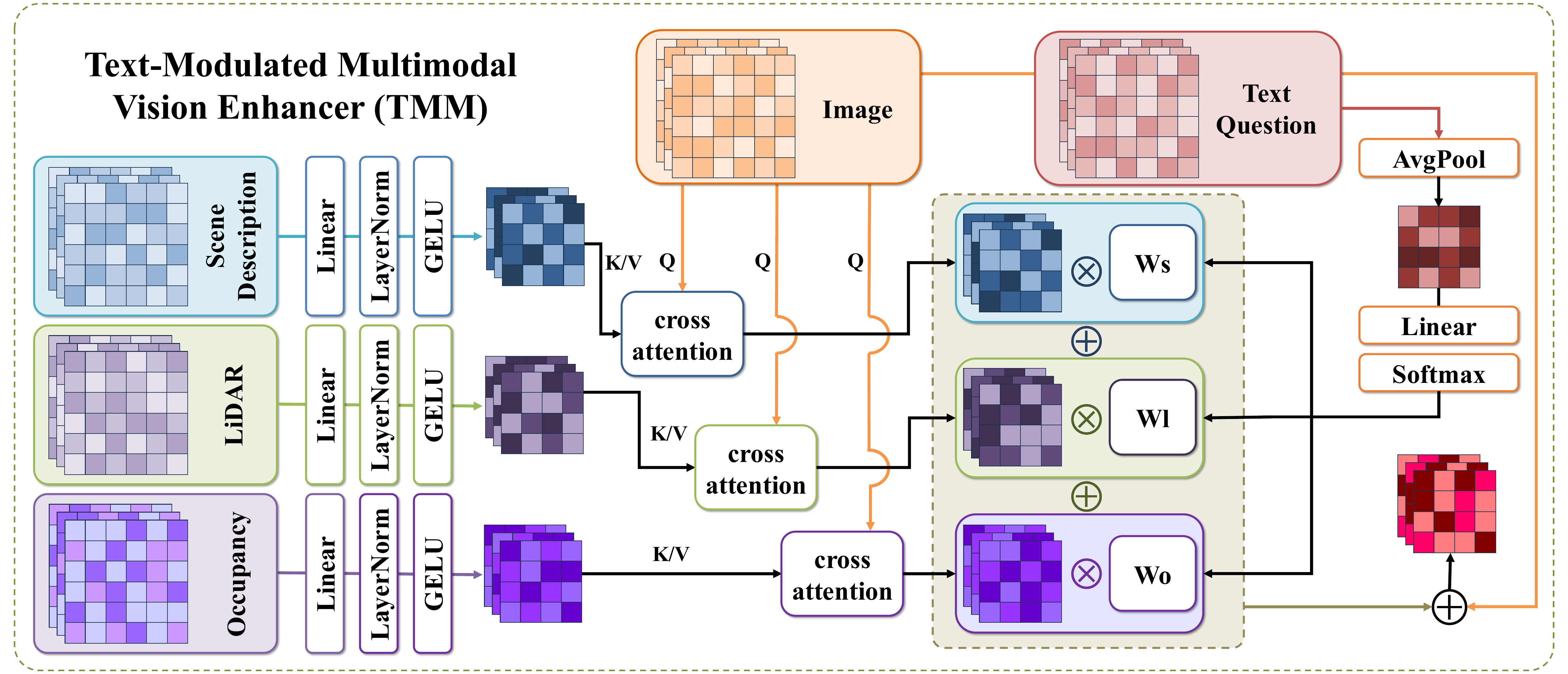}
     \end{center}
    % \vspace{-0.5cm}
 \caption{
% Architectural diagram of the Text-oriented Multimodal Modulator (TMM). The TMM achieves effective multimodal information fusion through three core steps. First, heterogeneous modal features are mapped to a unified latent space via linear projection. Next, adaptive fusion weights are generated based on the input text question features, dynamically adjusting the contribution of each modality during the fusion process. Finally, the TMM executes parallel cross-modal cross-attention mechanisms and performs weighted summation of multimodal information according to the predicted weights, thereby enhancing the scene representation capability of multimodal information.
Architectural diagram of the Text-oriented Multimodal Modulator (TMM). TMM achieves multimodal fusion via three steps: projecting multimodal features to latent space, generating text-question-driven adaptive fusion weights, and performing cross-modal cross-attention with weighted summation to enhance multimodal scene representation capability.
}
 \label{fig:tmm}
 \end{figure}

To meet the differentiated requirements of various driving scenario questions regarding multimodal information focus, this work proposes the TMM. The TMM employs question features as guidance signals and utilizes a learnable weight prediction network to adaptively aggregate multimodal information, achieving question-aware multimodal fusion, as illustrated in Figure~\ref{fig:tmm}. To enable subsequent multimodal fusion, occupancy, LiDAR, and scene description features are first mapped to a unified latent space dimension through linear projection, yielding $\tilde{\mathbf{F}}_{\mathrm{L}}$, $\tilde{\mathbf{F}}_{\mathrm{O}}$, and $\tilde{\mathbf{F}}_{\mathrm{D}}$, respectively. The TMM predicts the importance of each modality based on the semantic information of questions, thereby enhancing multimodal semantic alignment and promoting deep feature fusion. Given the text question features $\mathbf{F}_Q$, global average pooling is first applied:
\begin{equation}
\label{eq:question_pooling}
    \mathbf{F}^{G}_{Q} = \frac{1}{M} \sum_{m=1}^{M} \mathbf{F}_Q^{(m)},
\end{equation}
where $\mathbf{F}_Q^{(m)}$ denotes the feature representation of the $m$-th token in the question feature sequence.

Subsequently, a learnable weight predictor $\mathbf{W}_{\omega} \in \mathbb{R}^{3 \times D_m}$ is employed to generate weights for the three modalities: occupancy, LiDAR, and scene description:
\begin{equation}
\label{eq:weight_prediction}
    \boldsymbol{\omega} = \mathrm{Softmax}(\mathbf{W}_{\omega} \mathbf{f}_q^\top) \in \mathbb{R}^{B \times 3},
\end{equation}
where $\boldsymbol{\omega} = [\omega_{\mathrm{lidar}}, \omega_{\mathrm{occ}}, \omega_{\mathrm{desc}}]$, satisfying the normalization constraint $\sum_{m \in \{\mathrm{lidar}, \mathrm{occ}, \mathrm{text}\}} \omega_m = 1$ and $\omega_m \geq 0$. This design enables the model to adaptively allocate attention according to question requirements.

After obtaining the fusion weights, the TMM executes three parallel cross-modal cross-attention operations, using image features $\mathbf{F}_{\mathrm{img}}$ as queries, which interact with multimodal features as keys and values respectively, to achieve cross-modal information alignment and fusion:
\begin{equation}
 \left\{
 \begin{aligned}
 \label{eq:cross_attn_lidar}
     \mathbf{E}_{\mathrm{L}} &= \mathrm{Softmax}\!\left( \frac{ \mathbf{F}_{\mathrm{I}} \, \tilde{\mathbf{F}}_{\mathrm{L}}^{\top} }{\sqrt{d_k}} \right) \, \tilde{\mathbf{F}}_{\mathrm{L}}, \\
%  \label{eq:cross_attn_occ}
     \mathbf{E}_{\mathrm{O}} &= \mathrm{Softmax}\!\left( \frac{ \mathbf{F}_{\mathrm{I}} \, \tilde{\mathbf{F}}_{\mathrm{O}}^{\top} }{\sqrt{d_k}} \right) \, \tilde{\mathbf{F}}_{\mathrm{O}}, \\
%  \label{eq:cross_attn_text}
     \mathbf{E}_{\mathrm{D}} &= \mathrm{Softmax}\!\left( \frac{ \mathbf{F}_{\mathrm{I}} \, \tilde{\mathbf{F}}_{\mathrm{D}}^{\top} }{\sqrt{d_k}} \right) \, \tilde{\mathbf{F}}_{\mathrm{D}},
 \end{aligned}
 \right.
\end{equation}
where $d_k$ is the scaling factor.

The enhanced features $\mathbf{E}_M$ are combined through weighted summation using the predicted weights, followed by layer normalization:
 \begin{equation}
 \label{eq:weighted_fusion}
     \mathbf{E}_{\mathrm{fused}} = \mathbf{F}_\text{I} + \mathrm{LayerNorm}\left( \omega_{\mathrm{lidar}}\mathbf{E}_{\mathrm{lidar}} + \omega_{\mathrm{occ}}\mathbf{E}_{\mathrm{occ}} + \omega_{\mathrm{text}}\mathbf{E}_{\mathrm{text}} \right).
 \end{equation}
Finally, a residual connection strategy fuses image features with modulated multimodal features.

\subsection{Cross-Modal Abstractor}
\label{subsec:cma}

\begin{figure}
     \centering
     \includegraphics[width=0.95\linewidth]{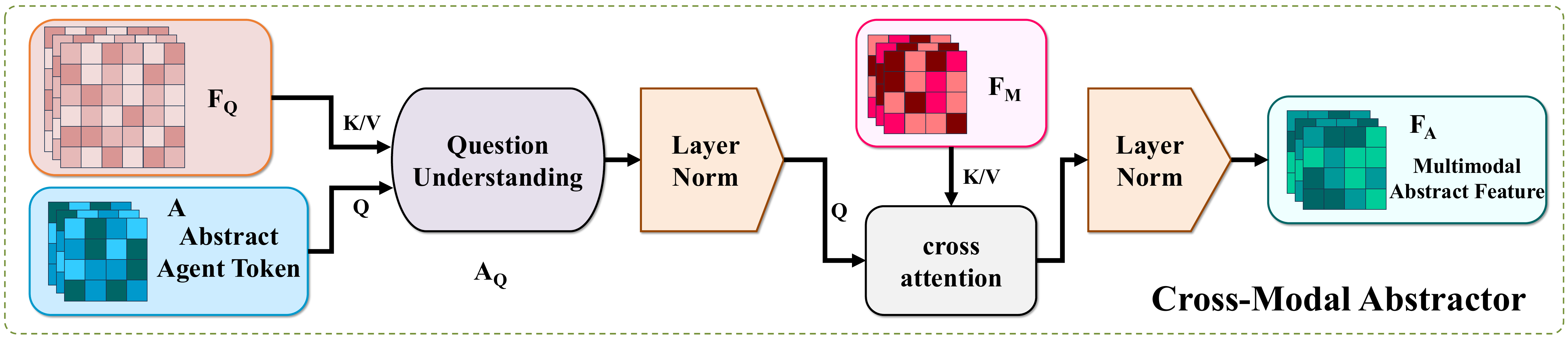}
     \caption{
     % Architectural diagram of the Cross-Modal Abstractor (CMA). The CMA employs a two-stage design to refine multimodal semantic information and generate abstractions. In the first stage, abstract tokens extract task requirements from question features to understand the semantic content of questions. In the second stage, through multimodal information refinement, the abstract tokens actively query multimodal features and summarize key information, thereby generating compact multimodal abstractions. This architecture effectively extracts and integrates task-relevant critical information, providing efficient and precise semantic support for subsequent reasoning.
     Architectural diagram of the Cross-Modal Abstractor (CMA). The CMA refines multimodal semantic information through a two-stage process: first, abstract tokens capture task-specific information from question features; then, multimodal features are queried and summarized to generate compact abstractions, enabling efficient reasoning support.
     }
     \label{fig:cma}
 \end{figure}

To enable the model to effectively extract key information from complex driving scenes, the CMA is introduced. The CMA refines multimodal semantic information through learnable abstract tokens $\mathbf{A}$, constructing abstractions that enable the LLM to efficiently focus on critical driving scene information, as illustrated in Figure~\ref{fig:cma}.

% \subsubsection{Stage I: Task Understanding via Question-Abstract Interaction}
The first stage aims to enable the learnable abstract tokens to understand the semantic information of the current question. Specifically, the abstract tokens serve as queries, while the text question features $\mathbf{F}_Q$ serve as both keys and values, extracting question semantic information through a multi-head cross-attention mechanism:
\begin{equation}
\label{eq:task_understanding}
    \mathbf{Q}_{\mathrm{A}} = \mathrm{Softmax}\!\left( \frac{ \mathbf{A} \, \mathbf{F}_Q^{\top} }{\sqrt{d_k}} \right) \, \mathbf{F}_Q.
\end{equation}
This is followed by residual connection and layer normalization operations. Through this process, $\mathbf{Q}_{\mathrm{A}}$ captures the core semantic information of the text question.

% \subsubsection{Stage II: Visual Evidence Extraction via Abstract-Vision Interaction}

Subsequently, the abstract tokens that have understood the text question serve as queries to query the fused multimodal features $\mathbf{F}_{\mathrm{enhanced}}$, extracting question-relevant scene information:
\begin{equation}
    \mathbf{F}_A = \mathrm{Softmax}\!\left( \frac{ \mathbf{A} \, \mathbf{F}_{\mathrm{M}} }{\sqrt{d_k}} \right) \, \mathbf{F}_{\mathrm{M}}.
\end{equation}
At this point, $\mathbf{F}_A$ contains a task-relevant compact visual abstraction, which is injected into the LLM, providing compact and semantically rich multimodal cues for the subsequent reasoning process. 
% Algorithm~\ref{alg:cma} summarizes the complete forward propagation procedure of the CMA.

\subsection{Large Language Model}
\label{subsec:llm_decoder}

After processing through TMM and CMA, \name{} constructs a unified multimodal input sequence. To clearly distinguish information from different modalities, special tokens are introduced into the input sequence, which is then fed into a pre-trained T5 model~\cite{raffel2020t5} for answer generation. This design enables the LLM to explicitly recognize modal boundaries and understand the logical progression from textual questions through cross-modal abstractions to visual scene representations. The LLM generates the answer sequence $S^{\text{GT}} = (s^{\text{GT}}_1, s^{\text{GT}}_2, \dots, s^{\text{GT}}_X)$ in an autoregressive manner, and the cross-entropy loss is computed as:
\begin{equation}
    \mathcal{L} = - \sum_{x=1}^{X} s^{\text{GT}}_x \log (\hat{s}_x),
\end{equation}
which measures the discrepancy between the predicted token sequence and the target sequence.

In summary, \name{} constructs a complete end-to-end autonomous driving VLM framework through multimodal information encoding, Text-oriented Multimodal Modulation (TMM), Cross-Modal Abstraction (CMA), and LLM-based decoding. The TMM dynamically adjusts the fusion weights of each modality based on question semantics, preventing information dilution. The CMA compresses cross-modal semantics through learnable abstract tokens, enabling the LLM to efficiently focus on critical cues. These two innovative modules work synergistically to achieve a paradigm shift from "image understanding" to "scene understanding", providing a robust and efficient solution for visual question answering tasks in complex driving scenarios.

\section{Experiments}
\label{sec:experiments}

\begin{table*}[!t]
\centering
\caption{Quantitative comparison on the DriveLM dataset. $\uparrow$ indicates that higher values are better, while $\downarrow$ denotes that lower values are better. \textbf{Bold} values indicate the best performance, and \underline{underline} indicate the second-best performance.}
% \vspace{-0.1cm}
\label{tab:drivelm_results}
\resizebox{0.98\textwidth}{!}{%
\begin{tabular}{lcc|cccc}
\toprule
\multirow{2}{*}{\textbf{Methods}} & \multirow{2}{*}{\textbf{Venues}} & \multirow{2}{*}{\textbf{Inference Schema}} & \multicolumn{4}{c}{\textbf{DriveLM}} \\
\cmidrule{4-7}
& & & \textbf{BLEU-4 $\uparrow$} & \textbf{METEOR $\uparrow$} & \textbf{ROUGE-L $\uparrow$} & \textbf{CIDEr $\uparrow$} \\
\midrule
DriveLM-Agent \cite{sima2024drivelm} & ECCV'24 & Graph & \underline{53.09} & 36.19 & 66.79 & 2.79  \\
EM-VLM4AD$_\text{Base}$ \cite{Gopalkrishnan2024emvlm4ad} & CVPR'24 & Single & 45.36 & 34.49 & 71.98 & 3.20 \\
EM-VLM4AD$_\text{Q-Large}$ \cite{Gopalkrishnan2024emvlm4ad} & CVPR'24 & Single & 40.11 & 34.34 & 70.72 & 3.10 \\
% InternVL-2 \cite{} & -- & Single & 51.42 & 37.12 & 77.08 & 3.53  \\
LLaMA-Adapter \cite{zhang2024llamaadapter} & ICLR'24 & Single & 45.96 & 33.66 & 69.78 & 3.07 \\
% MiniDrive$_{224}$ \cite{zhang2024minidrive} & arXiv'25 & Single & 49.70 & 36.30 & 73.30 & 3.28  \\
MiniDrive \cite{zhang2024minidrive} & arXiv'25 & Single & 50.20 & 37.40 & 73.50 & 3.32  \\
LaVida Drive \cite{jiao2024lavida} & arXiv'25 & Single & 51.30 & 38.00 & 73.90 & 3.32 \\
MPDrive \cite{zhang2025mpdrive} & CVPR'25 & Single & 52.71 & \underline{38.31} & \textbf{76.98} & \underline{3.56} \\

\midrule
\textbf{\bf{\name{}} (Ours)} & \textbf{-} & Single & \textbf{54.56} & \textbf{41.78} & \underline{75.27} & \textbf{3.63} \\
\bottomrule
\label{tab_drivelm}
\end{tabular}
}
\end{table*}

\subsection{Experiment Settings}

\paragraph{Datasets.}
Experiments are conducted on the DriveLM~\cite{sima2024drivelm} and NuScenes-QA~\cite{qian2024nuscenesqa} benchmarks, which are designed to evaluate VLMs in autonomous driving scenarios.
This study employs the DriveLM dataset, a multi-view VQA dataset designed for autonomous driving tasks. Derived from the nuScenes dataset, it encompasses core autonomous driving tasks, including perception, planning, and decision-making. Each sample consists of images from multiple viewpoints, accompanied by corresponding question-answer pairs, facilitating VLM tasks in autonomous driving scenarios. For fair comparison with baselines, the training and evaluation protocols of prior works~\cite{Gopalkrishnan2024emvlm4ad,zhang2024minidrive} are strictly followed.
NuScenes-QA is a multimodal vision-language question answering benchmark specifically designed for autonomous driving scenes. The dataset includes 34K visual scenes and 460K question-answer pairs, providing five question types: existence, counting, query-object, query-status, and comparison. For a fair comparison with baselines, the training and evaluation protocols from prior work~\cite{qian2024nuscenesqa} are strictly followed. NuScenes-QA provides a comprehensive and challenging benchmark for vision-language question answering tasks in autonomous driving scenes.

\subsubsection{Evaluation on NuScenes-QA}
\begin{table*}[!t]
\centering
\caption{Results of different models on the NuScenes-QA test set. \textbf{Bold} values indicate the best performance, and \underline{underline} indicate the second-best performance.}
\label{tab:nuscenes_qa}
\resizebox{\textwidth}{!}{%
\begin{tabular}{l|ccc|ccc|ccc|ccc|ccc|c}
\hline
\multicolumn{1}{c|}{\multirow{2}{*}{Models}} & \multicolumn{3}{c|}{Exist} & \multicolumn{3}{c|}{Count} & \multicolumn{3}{c|}{Object} & \multicolumn{3}{c|}{Status} & \multicolumn{3}{c|}{Comparison} & \multirow{2}{*}{Acc} \\ \cline{2-16} 
\multicolumn{1}{c|}{} & H0 & H1 & All & H0 & H1 & All & H0 & H1 & All & H0 & H1 & All & H0 & H1 & All &  \\ \hline
\multicolumn{1}{l|}{Q-Only~\cite{qian2024nuscenesqa}} & 81.7 & 77.9 & 79.6 & 17.8 & 16.5 & 17.2 & 59.4 & 38.9 & 42.0 & 57.2 & 48.3 & 51.3 & 79.5 & 65.7 & 66.9 & 53.4 \\ 
% \hline \hline
BEVDet+BUTD~\cite{qian2024nuscenesqa} & \underline{87.2} & 80.6 & 83.7 & 21.7 & 20.0 & 20.9 & 69.4 & 45.2 & 48.8 & 55.0 & 50.5 & 52.0 & 76.1 & 66.8 & 67.7 & 57.0 \\
CenterPoint+BUTD~\cite{qian2024nuscenesqa} & \textbf{87.7} & 81.1 & 84.1 & 21.9 & \underline{20.7} & \underline{21.3} & 70.2 & 45.6 & 49.2 & 62.8 & 52.4 & 55.9 & 81.6 & 68.0 & 69.2 & 58.1 \\
% MSMDFusion+BUTD & 89.4 & 81.4 & 85.1 & 25.3 & 21.3 & 23.2 & 73.3 & 48.7 & 52.3 & 67.4 & 55.4 & 59.5 & 81.6 & 67.2 & 68.5 & 59.8 \\
% GroundTruth+BUTD & 98.9 & 87.2 & 92.6 & 76.8 & 38.7 & 57.5 & 99.7 & 71.9 & 76.0 & 98.8 & 81.9 & 87.6 & 98.1 & 76.1 & 78.1 & 79.2 \\ 
% \hline
BEVDet+MCAN~\cite{qian2024nuscenesqa} & \underline{87.2} & \underline{81.7} & \underline{84.2} & 21.8 & 19.2 & 20.4 & \textbf{73.0} & 47.4 & 51.2 & 64.1 & 49.9 & 54.7 & 75.1 & 66.7 & 67.4 & 57.9 \\
CenterPoint+MCAN~\cite{qian2024nuscenesqa} & \textbf{87.7} & \textbf{82.3} & \textbf{84.8} & \underline{22.5} & 19.1 & 20.8 & 71.3 & \underline{49.0} & \underline{52.3} & \underline{66.6} & \underline{56.3} & \underline{59.8} & \underline{82.4} & \underline{68.8} & \underline{70.0} & \underline{59.5} \\
% MSMDFusion+MCAN & 89.0 & 82.3 & 85.4 & 23.4 & 21.1 & 22.2 & 75.3 & 50.6 & 54.3 & 69.0 & 56.2 & 60.6 & 78.8 & 68.8 & 69.7 & 60.4 \\
% GroundTruth+MCAN & 99.6 & 95.5 & 97.4 & 52.7 & 39.9 & 46.2 & 99.7 & 86.2 & 88.2 & 99.3 & 95.4 & 96.8 & 99.7 & 90.2 & 91.0 & 84.3 \\ 
\hline
% \name{}(Ours) & 85.7 & 80.8 & 83.1 & \textbf{26.7} & \textbf{28.2} & \textbf{27.5} & \underline{72.4} & \underline{48.4} & \underline{51.9} & \textbf{66.8} & \textbf{58.5} & \textbf{61.3} & \textbf{84.7} & \textbf{72.5} & \textbf{73.6} & \textbf{61.0} \\
\name{}(Ours) & 86.7 & 81.6 & 83.9 & \textbf{28.1} & \textbf{30.3} & \textbf{29.2} & \underline{72.1} & \textbf{51.2} & \textbf{54.3} & \textbf{69.3} & \textbf{60.5} & \textbf{63.5} & \textbf{85.4} & \textbf{74.4} & \textbf{75.3} & \textbf{62.7} \\
\hline
\end{tabular}%
}
\end{table*}

% ablation study on multimodal fusion
\begin{table}[htbp]
    \centering
    \caption{%Comparison results on representative fusion strategies.
    Comparison of multimodal fusion strategies on the DriveLM benchmark. Incorporating Occupancy (O), LiDAR (L), and textual scene description (T) progressively improves performance, with the full multimodal configuration achieving the best results across all metrics. $\uparrow$ indicates that higher values are better. \textbf{Bold} indicates the highest value.}
    % \vspace{-0.1cm}
    \label{tab:ablation_study_1}
    \resizebox{0.6\columnwidth}{!}{
    \begin{tabular}{l|cccc}
    \toprule
    \multirow{2}{*}{\textbf{Modalities}}  & \multicolumn{4}{c}{\textbf{DriveLM}} \\
    \cmidrule{2-5}
    & \textbf{BLEU-4 $\uparrow$} & \textbf{METEOR $\uparrow$} & \textbf{ROUGE-L $\uparrow$} & \textbf{CIDEr $\uparrow$} \\
    \midrule
    -- & 50.93 & 37.38 & 71.24 & 3.12 \\
    L & 51.91 & 39.38 & 72.33 & 3.25 \\
    L+T & 52.87 & 40.18 & 73.33 & 3.39 \\
    L+T+O & \textbf{54.56} & \textbf{41.78} & \textbf{75.27} & \textbf{3.63} \\
    \bottomrule
    \end{tabular}
    }
    % \\[0.1cm]
    % \parbox{0.98\columnwidth}{\footnotesize $\uparrow$ indicates that higher values are better. \textbf{Bold} indicates the highest value.}
\end{table}

\paragraph{Implementation Details.}
The experiment utilizes 8 A100 GPUs for synchronized training, with a batch size of 128. The language model employs T5 as both the encoder and LLM decoder, with a default maximum text length of 500 and a sequence length limit of 512. The vision encoder utilizes UniRepLKNet, with Occupancy feature encoding performed using UniScene and LiDAR encoding based on Micheal. All encoder parameters are frozen during training. To enhance efficiency, we pre-encode the features of Occupancy, LiDAR, and Scene Descriptions, which are loaded during training. The model is optimized using the AdamW optimizer, with an initial learning rate of $1 \times 10^{-4}$ and a weight decay coefficient of 0.01. The learning rate scheduling follows a cosine annealing strategy.

\paragraph{Metrics.}
To ensure a fair and accurate evaluation of model performance on the DriveLM dataset, we follow the evaluation protocols established in prior works~\cite{Gopalkrishnan2024emvlm4ad,zhang2024minidrive} and employ widely used natural language generation metrics: BLEU‑4, METEOR, ROUGE‑L, and CIDEr. Specifically:
BLEU-4 measures the precision of 4-gram matching between the prediction and reference texts. METEOR accounts for synonym and stem matching to better capture semantic similarity. ROUGE-L focuses on assessing the length of the longest common subsequence. CIDEr evaluates consistency using TF-IDF weighted n-gram matching.

NuScenes-QA uses Top-1 accuracy as the evaluation metric to assess model performance on the overall test set and across different question types~\cite{qian2024nuscenesqa}. The question types are primarily divided into five categories: 1) Existence: querying whether a specific object exists in the scene; 2) Count: counting objects under specific conditions; 3) Object: recognizing objects based on linguistic descriptions; 4) Status: querying the state of a specified object; 5) Comparison: comparing specified objects or their states. These questions are further divided into zero-hop (H0) and one-hop (H1) categories, representing simpler vision-based reasoning tasks and those requiring reasoning about object relationships, respectively.

\subsection{Comparison With State-of-the-Art Methods }
\subsubsection{Evaluation on DriveLM}

\name{} is evaluated on the DriveLM benchmark and compared with several representative baselines. The results are shown in Table 1, which include the graph-based reasoning approach DriveLM-Agent, multi-view fusion models like EM-VLM4AD and MiniDrive, and recent state-of-the-art approaches  such as LaVida Drive and MPDrive.
The experimental results are reported in Table~\ref{tab_drivelm}. \name{} outperforms the second-best DriveLM-Agent by 1.47 in the BLEU-4 metric, demonstrating a significant advantage in n-gram matching accuracy. This highlights the method's ability to generate content that more precisely matches the reference text.
In the METEOR metric, \name{} achieves a score of 41.78, surpassing the second-best MPDrive by 3.47, further demonstrating the method's advantage in semantic consistency and similarity to the reference text. Compared to other methods, \name{} shows significant improvements in both precision and semantic matching.
In the ROUGE-L metric, \name{} performs comparably to MPDrive, demonstrating its strong competitiveness in long-sequence matching and sentence structure coverage.
Moreover, \name{} surpasses all comparison methods in the CIDEr metric, further consolidating its advantage in text generation consistency and semantic quality.
Overall, \name{} achieves optimal performance across multiple metrics, demonstrating significant advantages over existing methods in accuracy, semantic consistency, long-sequence matching, and generation consistency. The experimental results thoroughly validate the superiority and effectiveness of \name{}.

Multiple baselines are evaluated on the NuScenes-QA dataset, which combine advanced 3D detection methods with vision language question answering frameworks for feature extraction and question answering. As shown in Table~\ref{tab:nuscenes_qa}, \name{} significantly outperforms the baselines across multiple tasks, particularly in count, status, and comparison, demonstrating its strong reasoning capabilities in autonomous driving scenarios. The superiority of \name{} stems from its efficient multimodal fusion mechanism, which integrates occupancy grids, LiDAR, and scene description information, thereby enhancing its reasoning capabilities in complex environments. Its performance improvements across multiple dimensions demonstrate the reliability and advanced nature of \name{} as an autonomous driving VLM.

% ablation study on TMM and CMA
\begin{table}[htbp]
    \centering
    \caption{Ablation study results on the DriveLM dataset. The experiments compare four model configurations. Bold values indicate the best performance. $\uparrow$ indicates that higher values are better. \textbf{Bold} indicates the highest value.}
    % \vspace{-0.1cm}
    \label{tab:ablation_study_2}
    \resizebox{0.6\columnwidth}{!}{%
    \begin{tabular}{cc|cccc}
    \toprule
    \multirow{2}{*}{\textbf{TMM}} & \multirow{2}{*}{\textbf{CMA}} & \multicolumn{4}{c}{\textbf{DriveLM}} \\
    \cmidrule{3-6}
    & & \textbf{BLEU-4 $\uparrow$} & \textbf{METEOR $\uparrow$} & \textbf{ROUGE-L $\uparrow$} & \textbf{CIDEr $\uparrow$} \\
    \midrule
    -- & -- & 49.89 & 36.50 & 70.65 & 3.07 \\
    $\checkmark$ & -- & 52.72 & 39.21 & 73.12 & 3.25 \\
    -- & $\checkmark$ & 50.93 & 37.38 & 71.24 & 3.12 \\
    $\checkmark$ & $\checkmark$ & \textbf{54.56} & \textbf{41.78} & \textbf{75.27} & \textbf{3.63} \\
    \bottomrule
    \end{tabular}
    }
    \\[0.1cm]
    % \parbox{0.98\columnwidth}{\footnotesize $\uparrow$ indicates that higher values are better. \textbf{Bold} indicates the highest value.}
\end{table}

% ablation study on number of agent tokens
\begin{table}[htbp]
    \centering
    \caption{Ablation study results on the number of agent tokens in the CMA module. $\uparrow$ indicates that higher values are better. \textbf{Bold} indicates the highest value.}
    % \vspace{-0.1cm}
    \label{tab:ablation_study_3}
    \resizebox{0.6\columnwidth}{!}{%
    \begin{tabular}{c|cccc}
    \toprule
    \multirow{2}{*}{\textbf{Tokens}}  & \multicolumn{4}{c}{\textbf{DriveLM}} \\
    \cmidrule{2-5}
    & \textbf{BLEU-4 $\uparrow$} & \textbf{METEOR $\uparrow$} & \textbf{ROUGE-L $\uparrow$} & \textbf{CIDEr $\uparrow$} \\
    \midrule
    8 & 52.95 & 40.12 & 73.00 & 3.47 \\
    16 & \textbf{54.56} & \textbf{41.78} & \textbf{75.27} & \textbf{3.63} \\
    24 & 53.99 & 41.75 & 74.68 & 3.43 \\
    32 & 53.07 & 41.99 & 74.53 & 3.41 \\
    \bottomrule
    \end{tabular}
    }
    \label{tab:token_num}
    \\[0.1cm]
    % \parbox{0.98\columnwidth}{\footnotesize $\uparrow$ indicates that higher values are better. \textbf{Bold} indicates the highest value.}
\end{table}

\subsection{Ablation Studies}

To analyze the contribution of different modalities to model performance, we conduct multimodal ablation experiments on the DriveLM dataset, as presented in Table~\ref{tab:ablation_study_1}. The model using only image information (Image-Only) serves as the baseline, showing relatively low overall performance, which indicates that 2D visual cues alone are insufficient for fine-grained understanding in complex driving scenarios.
Incorporating the LiDAR modality (L) leads to consistent improvements across all metrics, demonstrating the complementary value of point cloud depth cues in strengthening scene understanding. Adding the the generated scene-level textual descriptions (L+T) further enhances performance, indicating that high-level semantic text provides additional semantic constraints and reasoning cues for visual and geometric features. Finally, integrating the Occupancy modality (L+T+O) yields the best performance across all metrics, showing that dense 3D occupancy information effectively complements spatial structure and layout cues, mitigating the limitations spatial perception. 
Overall, multimodal fusion enhances the model’s perception and understanding across spatial structure, depth geometry, and high-level semantics, thereby validating the effectiveness and necessity of the proposed multimodal design for scene understanding in autonomous driving.

% ----------------------------------------------

To validate the effectiveness of the proposed modules, we conduct systematic ablation experiments on the DriveLM benchmark, as shown in Table~\ref{tab:ablation_study_2}. Three model variants are considered: a baseline configuration without TMM or CMA, a variant incorporating only TMM, and the full model integrating both TMM and CMA.
The experimental results demonstrate that incorporating TMM leads to consistent improvements across multiple evaluation metrics, validating its effectiveness in dynamically modulating and integrating multimodal information. 
Furthermore, when CMA is added, the model performance improves further, indicating that CMA plays a key role in enhancing environment understanding and generation quality. 
Ultimately, the joint application of TMM and CMA achieves significant synergistic benefits, enabling the model to achieve superior overall performance in autonomous driving scene understanding tasks.

% figure_scenedesc
\begin{figure}
    \begin{center}
        \includegraphics[width=0.95\linewidth]{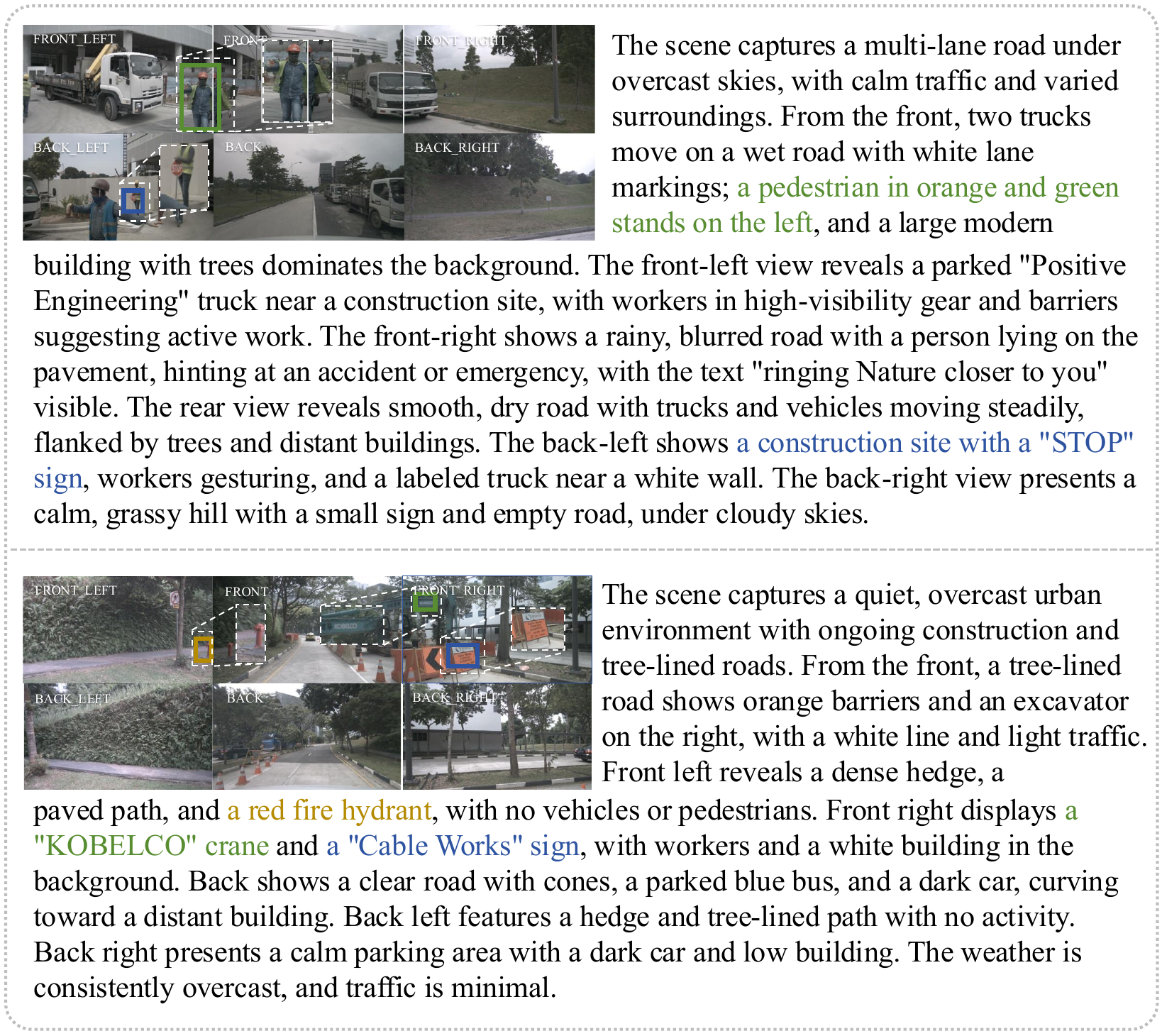}
    \end{center}
   % \vspace{-0.5cm}
\caption{Scene descriptions generated by a hierarchical two-stage method, highlighting accurate identification of objects and activities in complex scenarios.
}
\label{fig:sd_samples}
\end{figure}

% ----------------------------------------------

To analyze the impact of different agent token quantities in the CMA module on model performance, we conduct a sensitivity analysis, as shown in Table~\ref{tab:ablation_study_3}.
The experimental results demonstrate a nonlinear relationship between model performance and the number of tokens. 
When fewer tokens are used, the model fails to adequately express semantic information, limiting performance.
With a moderate token configuration, all metrics achieve optimal performance, indicating that the model strikes a favorable balance between expressiveness and redundancy control.
Further increasing the number of  tokens introduces redundant information and reduces the concentration of attention, leading to performance degradation.
This trend aligns with previous observations~\cite{Dai23InstructBLIP}, where a moderate number of tokens achieves an optimal trade-off between expressiveness and compactness
Based on the results in Table~\ref{tab:token_num}, we set the number of agent tokens to 16 to achieve the best overall performance.

\subsection{Qualitative Results}

\subsubsection{Scene Descriptions Visualization} 
This paper presents scene textual descriptions generated using a two-stage strategy. The approach generates high-quality textual descriptions from multi-view images, accurately capturing difficult to recognize objects and information, as shown in Figure~\ref{fig:sd_samples}.
For example, in the first case, the ``STOP'' sign in the left rear view overlaps with a pedestrian, making object recognition difficult. However, the scene description generated by \name{} successfully identifies and accurately describes the target. In the second case, \name{} not only accurately identifies small objects such as the red fire hydrant but also successfully detects non-traditional objects, such as a crane. Additionally, it provides an accurate description of a label that is difficult to discern with the naked eye, as its color closely resembles the surrounding barriers. Traditional methods struggle to capture such details, whereas \name{} provides a clear textual description, further demonstrating its robust recognition and description capabilities in complex scenes.
These results demonstrate that MMDrive significantly enhances scene understanding and description capabilities in autonomous driving tasks. It accurately captures and describes hard-to-recognize objects and complex scenarios, providing high-quality scene text descriptions and semantic cues, thereby offering strong support for vision-language tasks.

% figure_vis1
\begin{figure}
    \begin{center}
        \includegraphics[width=0.95\linewidth]{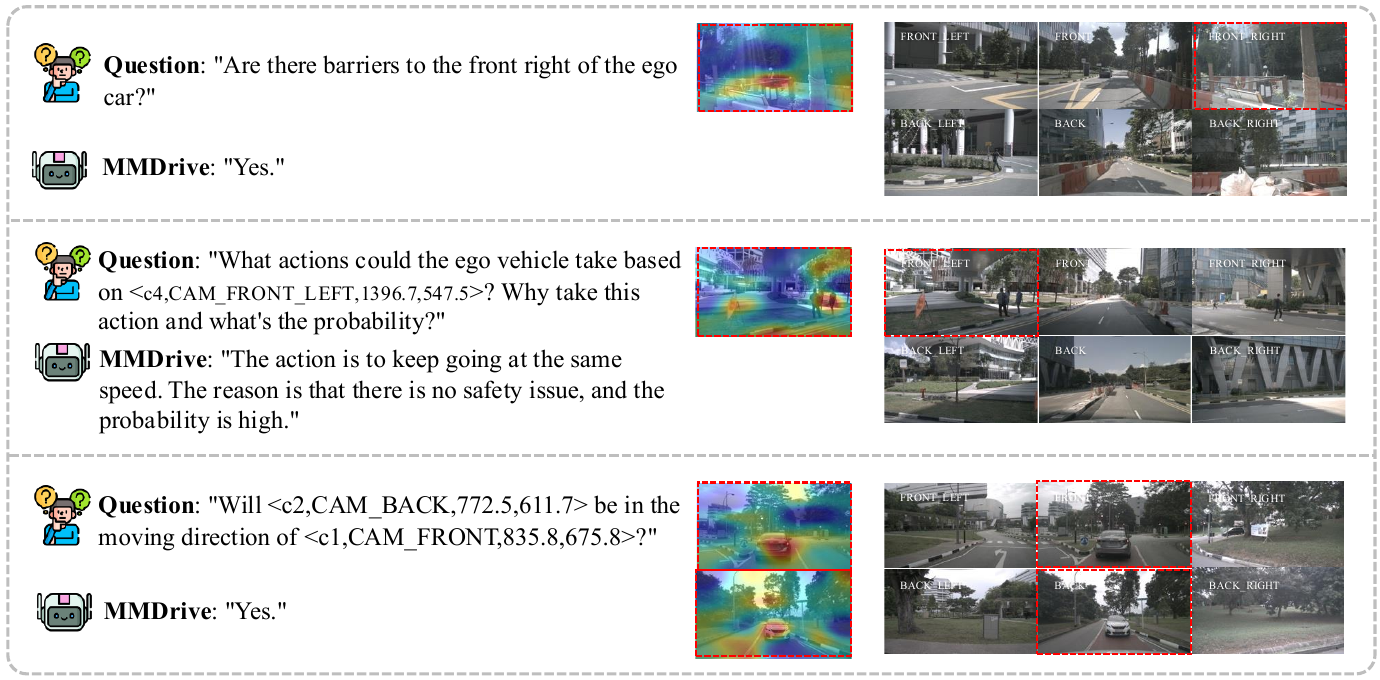}
    \end{center}
   % \vspace{-0.5cm}
\caption{Qualitative results of \name{} on multi-modal scene understanding. The left panel shows the questions and corresponding predictions, while the right panel displays six surround-view images. \name{} accurately detects objects, recommends safe behaviors, and predicts object details across different viewpoints.
}
\label{fig:vis1}
\end{figure}
\subsubsection{VLM Visualization} 
The qualitative results of \name{} are presented to evaluate its performance on multi-modal understanding tasks. \name{} demonstrates outstanding performance in multi-modal scene understanding tasks. As shown in Figure~\ref{fig:vis1}, the left panel lists the specific questions and the corresponding predictions generated by \name{}, while the right panel displays the six surround-view images related to each question.
It can be observed that \name{} consistently generates highly accurate answers. In the perception task, the system successfully identifies barriers. In the behavior task, the model provides accurate and safe behavior recommendations. In the prediction task, \name{} accurately determines the driving direction of objects across different viewpoint images. These results demonstrate that \name{} possesses strong multi-modal understanding and reasoning capabilities in complex scenarios, providing reliable support for VLM in autonomous driving tasks.

% figure_vis_hard
\begin{figure}
    \begin{center}
        \includegraphics[width=0.95\linewidth]{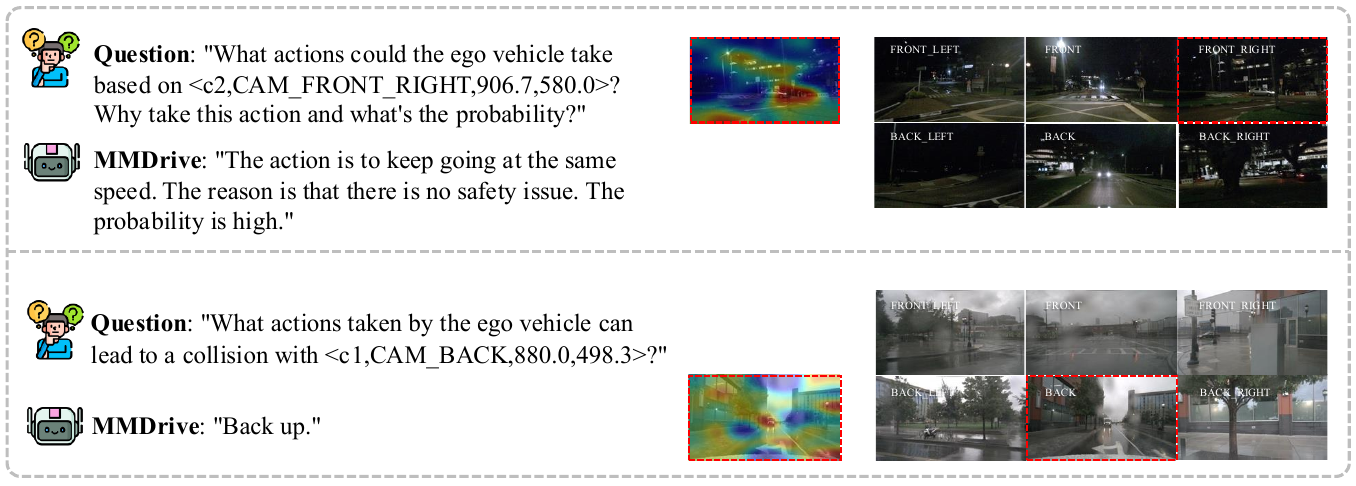}
    \end{center}
   % \vspace{-0.5cm}
\caption{Qualitative results of \name{} in challenging scenarios. The left panel shows the questions and corresponding predictions, while the right panel displays the six surround-view images. \name{} accurately answers questions under difficult visual conditions.
}
\label{fig:vis_hard}
\end{figure}
\begin{figure}
    \begin{center}
        \includegraphics[width=0.95\linewidth]{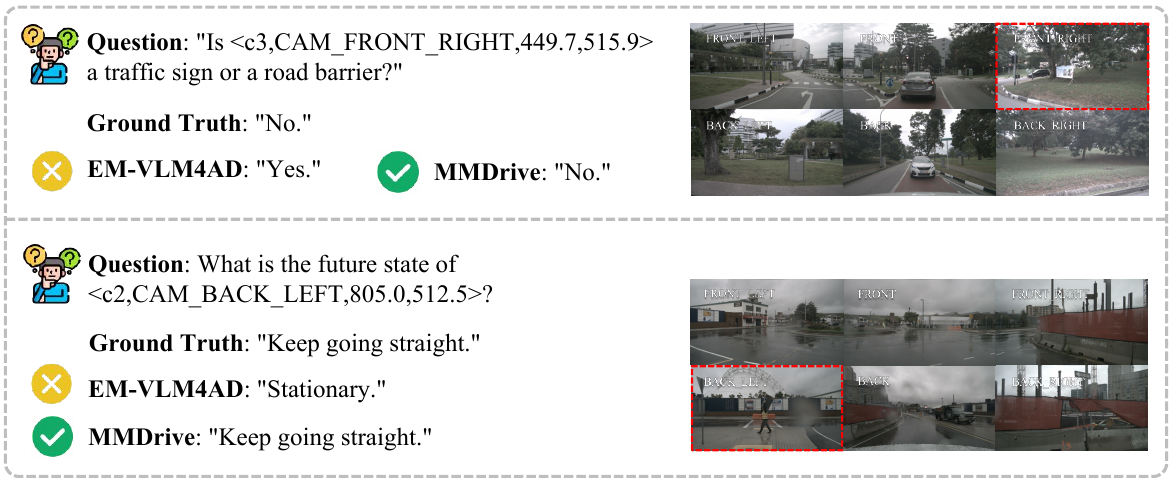}
    \end{center}
   % \vspace{-0.5cm}
\caption{Qualitative comparison between \name{} and EM-VLM4AD. \name{} outperforms EM-VLM4AD in prediction accuracy and reliability.
}
\label{fig:vis_compare}
\end{figure}

%
% \paragraph{Challenging Scenarios.} 
Figure~\ref{fig:vis_hard} illustrates the qualitative results of \name{} in challenging scenarios, such as nighttime and rainy conditions. For instance, in the nighttime scenario, despite poor visibility, the system accurately determines the vehicle's actions. In the rainy scenario, even with camera obstruction due to rain, \name{} still provides accurate predictions. These results highlight \name{}'s robust performance in multi-modal understanding tasks, maintaining high accuracy in complex environments. Moreover, Figure~\ref{fig:vis_compare} presents a qualitative comparison between \name{} and EM-VLM4AD. These results demonstrate that \name{}'s predictions are closer to the ground truth, highlighting its significant advantage in reliability and robustness compared to the baseline model.

% figure_vis_compare

\section{Conclusions}
\label{sec:conclusions}

In this work, an end-to-end VLM, \name{}, has been proposed to enhance multimodal scene understanding in autonomous driving. 
The framework extends traditional ``image understanding'' to generalized ``scene understanding'' by integrating three complementary modalities: occupancy grids, LiDAR point clouds, and scene descriptions. 
\name{} incorporates two synergistic innovation modules: TMM and CMA, which enable adaptive cross-modal fusion and efficient key information extraction through dynamic weight adjustment and compact cross-modal summaries.
Comprehensive evaluation shows that \name{} excels across multiple evaluation metrics, demonstrating the effectiveness and superiority of the proposed method.
Future extensions of \name{} will focus on its application in multimodal reasoning tasks such as long-term prediction, collaborative planning, and interpretable decision generation. Research will also explore lightweight deployment strategies to enable seamless integration of this framework into practical autonomous driving systems. Overall, this study provides a solid foundation for multimodal vision-language modeling in autonomous driving systems.

\printcredits

%% Loading bibliography style file
% \bibliographystyle{model1-num-names}
% \bibliographystyle{cas-model2-names}
\bibliographystyle{model1-num-names}
\providecommand{\URLprefix}{}

% Loading bibliography database
\bibliography{cas-refs}

\end{document}